\documentclass{article}

\usepackage[preprint]{nips_2018}

\usepackage[utf8]{inputenc} 
\usepackage[T1]{fontenc}    
\usepackage{hyperref}       
\usepackage{url}            
\usepackage{booktabs}       
\usepackage{amsfonts}       
\usepackage{nicefrac}       
\usepackage{microtype}      
\usepackage{graphicx, wrapfig, subcaption, setspace}
\usepackage[nolist]{acronym}
\usepackage{textcomp}
\usepackage{amsmath}

\graphicspath{{figures/}}   

\title{Object detection for crabs in top-view seabed imagery}

\author{
  Vlad Velici \\
  School of Electronics and Computer Science\\
  University of Southampton\\
  Southampton, UK
  \And
  Adam Pr\"ugel-Bennett \\
  School of Electronics and Computer Science \\
  University of Southampton \\
  Southampton, UK
}

\begin{document}

\maketitle

\begin{abstract}
  This report presents the application object detection on a database of underwater images of different species of crabs, as well as aerial images of sea lions and Pascal VOC. The model is an end-to-end object detection model based on a convolutional network and a Long Short-Term Memory detector.
\end{abstract}

\begin{acronym}
  \acro{mAP}[mAP]{mean average precision}
  \acro{cnn}[CNN]{Convolutional Neural Network}
  \acro{rpn}[RPN]{Region Proposal Network}
  \acro{rnn}[RNN]{Recurrent Neural Network}
  \acro{RNN}[RNN]{Recurrent Neural Network}
  \acro{LSTM}[LSTM]{Long Short-Term Memory}
  \acro{NLP}[NLP]{Natural Language Processing}
  \acro{GRU}[GRU]{Gated Recurrent Unit}
\end{acronym}

\section{Introduction} \label{section: tbx lit}

This report presents the problem of object detection and classification, popular datasets, and in Section~\ref{subsection: tbx bg-datasets} we present the datasets relevant to this work.

Image classification is the task of labelling an image with a class. Given an input image a model must predict what class it belongs to. Images used for classification often have one large central object. However this is not the case in real life where we are surrounded by many objects. A more challenging task is to predict where all the objects are in an image and what class they belong to. We call this object detection. An intermediary objective can be object localisation which is simply the task of finding all objects in an image but not assigning a class to them.

Depending on the actual dataset and desired end result, an object is defined by its class and either its $x$ and $y$ coordinates or a bounding box ($x, y$ coordinates, width $w$ and height $h$). We will use the term object coordinates to refer to either $x$ and $y$ coordinates or a bounding box.

Datasets like PASCAL Visual Object Classes (object detection challenge) \citep{Everingham15} and COCO \citep{lin2014microsoft} have bounding box coordinates, where other datasets like the ones used in the NOAA Fisheries Steller Sea Lion Population Count Kaggle competition \citep{noaakaggle} and the data obtained from \citep{thornton2016biometric} only have the centre coordinate of the objects. Semantic segmentation is a type of object detection where the bounding boxes or coordinates are replaced with object boundaries at pixel level. We are not looking into semantic segmentation in this report at this time.

\subsection{Crabs and Steller Sea Lions datasets} \label{subsection: tbx bg-datasets}

\begin{figure}
	\centering
	\includegraphics[width=0.9\textwidth]{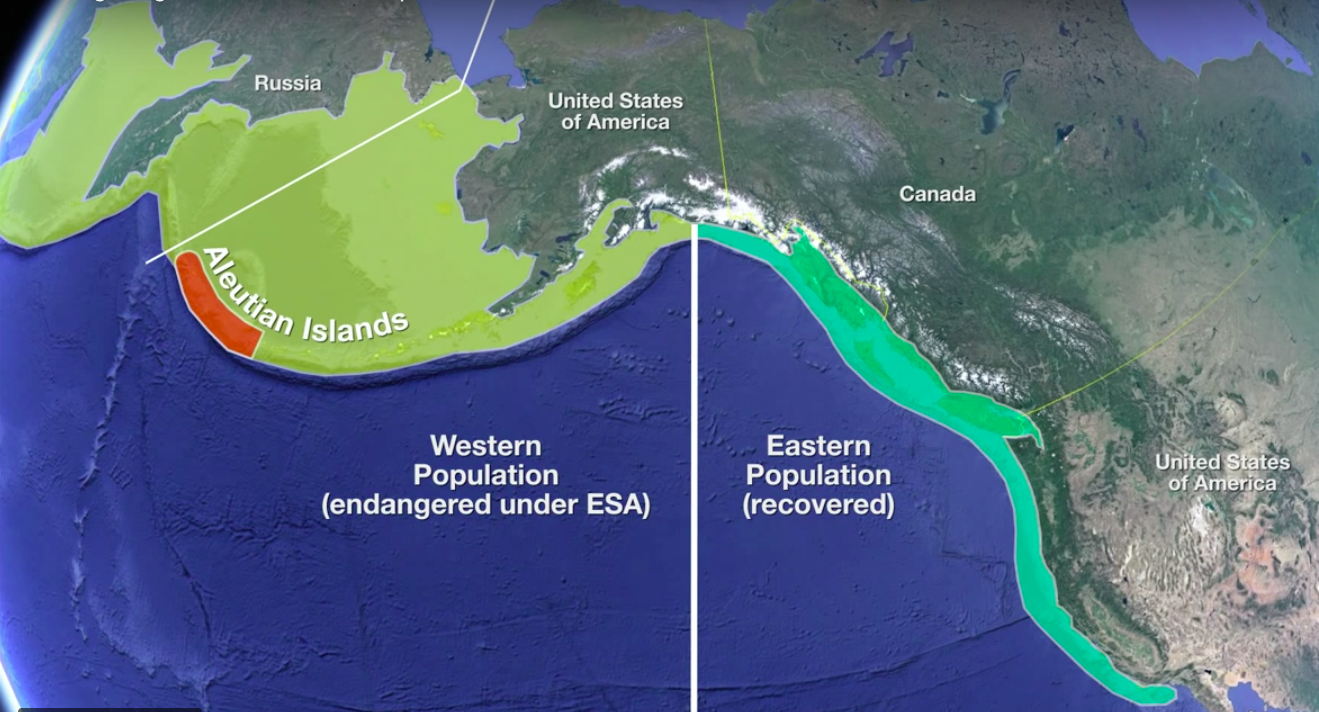}
	\caption[Map of Steller Sea Lion populations]{Map of Steller Sea Lion populations from NOAA Fisheries outreach video \url{https://youtu.be/oiL8tDCqzy4}. Red area shows the endangered population, yellow area shows the population that is increasing and green area shows the population that has fully recovered in 2013.}
	\label{fig:sealionmap}
\end{figure}

Steller Sea Lion population started to decline in the 1970s, and in 1990 it was listed as a threatened species. NOAA Fisheries divided the population into the western stock and the eastern stock, the separation line being at $144\deg$ West longitude. The eastern stock population started to recover in the late 1970s and has fully recovered in 2013. However, the western stock population has not. The population in the eastern side of the western stock ($144\deg$ - $170\deg$  West longitude) is increasing, but the western end, in Aleutian Islands, continues to decline. A map of the locations can be seen in Figure \ref{fig:sealionmap}. NOAA Fisheries is working on creating long-term population trends over time to help develop recovering strategies for the endangered population. Having such trends and tools will allow NOAA to make better decisions about managing fisheries and to identify possible threats affecting the endangered populations \citep{fritz2016aerial}.

NOAA Fisheries can now gather large amounts of aerial imagery efficiently using drones. They are, however, facing the problem of manually labelling the images to be able to create population trends. Currently they use human labelling but this method is prone to errors and slow, thus they are seeking automated solutions to improve the speed and quality of labelling. They have launched a Kaggle\footnote{Kaggle: \url{https://kaggle.com/}} competition to invite data scientists to find solutions and implement models to solve the labelling problem. The competition has a dataset of 100+GB of labelled imagery \citep{noaakaggle}.

The work presented in \cite{thornton2016biometric} is a practical method of using underwater robots to survey large areas of seafloor (multi-hectare areas). The images produced were then processed to build 3D reconstructions and mosaics which were manually labelled for 6 taxa of animals (crabs, mussels and shrimps). Labelling taxa by hand is a long, slow, expensive and error prone process, and it is the bottleneck of the whole process of creating population densities and distributions efficiently. They produced a dataset of two labelled mosaics of imagery taken at two locations: C0014G\_2m\_2014 is a 1060m deep drill site in the Iheya North field, and NBC\_2m\_2014 is a naturally active site located ~500m away.

In this report we will present a model created for these two datasets and report its performance, training and possible improvements.

\subsubsection{Note on the state of the art}

This work was part of my PhD \emph{initial} project, which was done in 2016, for the purpose of exploring deep learning and object detection, not necessarily to obtain good results. If the reader is interested in exploring the field of object detection and building competitive models we suggest to look up the current state of the art models and methods. We refrain from mentioning any works here because they will quickly become out of date.

\section{Crab detector network} \label{section: tbx}

In this report we present a deep neural network built for end-to-end object detection. As an end-to-end model, it takes as an input an image and outputs object coordinates, classes and confidence scores. The model can be configured to use centre coordinates of the objects or bounding boxes.

The model is built for the crabs and sea lions datasets where objects are defined by centre coordinates and are relatively small but it is also evaluated on VOC12 where objects are described by bounding boxes.

\section{Model architecture}

\begin{figure}
	\centering
	\includegraphics[width=\textwidth]{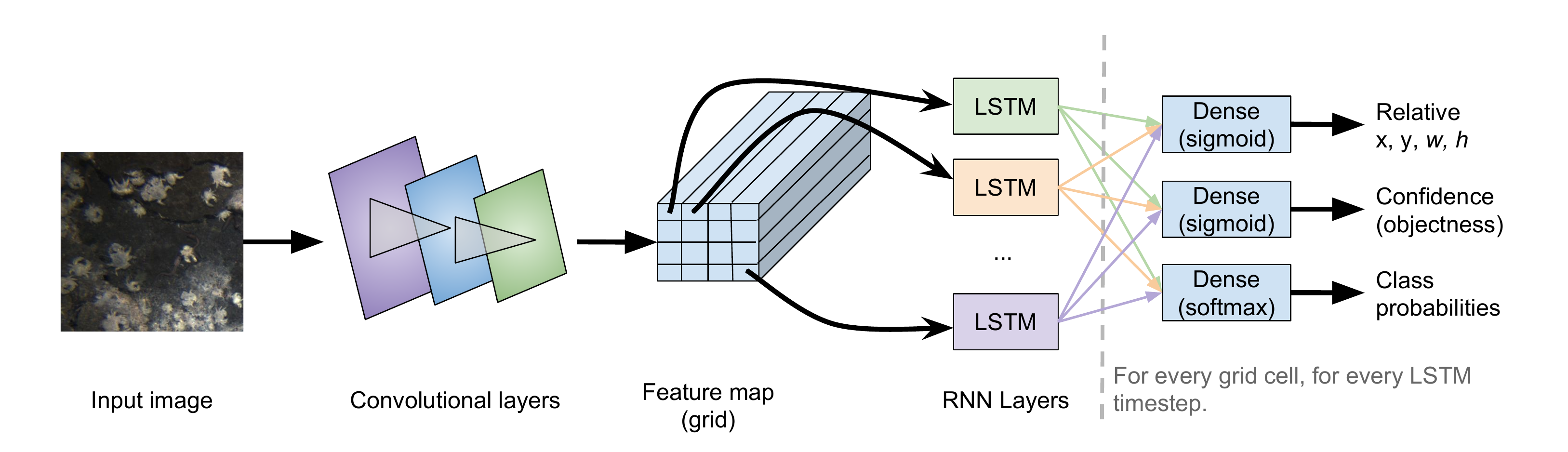}
	\caption{Architecture of our model.}
	\label{fig:model-diagram}
\end{figure}

The model architecture is based on \cite{stewart2016end}. We improved on their model by adding a classifier and adapting the loss function. We also added support for x, y coordinates labels as well as bounding boxes to enable using the model on the Pascal VOC 2012 dataset, the crabs and the Steller Sea Lions datasets.

The model architecture is illustrated in Figure \ref{fig:model-diagram}. For a forward pass, the model performs the following steps:

\textbf{Step 1} The input image is fed into a \ac{cnn}. The feature map that results from the \ac{cnn} is a $B \times W \times H \times F$ grid, where $B$ is the batch size, $W, H$ are the width and height of the grid and $F$ is the number of filters in the last convolutional layer.

\textbf{Step 2} Each of the $B \times W \times H$ grid cells ($F$-sized feature vectors) is fed into a layer of Long Short-Term Memory cells (LSTMs). Note that changes in $W$ and $H$ can be considered changes in the batch size $B$, therefore the model can be used for any input image size.

\textbf{Step 3} At each recurrence the LSTM cells are densely connected to three different units: one to predict objectness (used as confidence score), one to predict class probabilities (softmax), and one to predict object coordinates relative to the centre of the grid cell.

For our experiments we use square input images of $224 \times 224$ pixels, therefore our grid is a square of size $G = W = H$. We use two \ac{cnn} configurations that result in $G=7$ and $G=4$, respectively. Although we use a fixed image size as input, the model can support input images of any size as the LSTM and dense layers receive a fixed-size input regardless of the input image size. The grid size changes with the input image and affects how many times the LSTM and dense layers are run sequentially across the image.

A fixed number of recurrences (time steps) of the LSTM cells, $k$, is set. This is an implementation limitation. Theoretically, at test time the model can run recurrences until a stop symbol is produced (first object with confidence below a set threshold, e.g. $0.5$). During training the model can have one recurrence for each ground truth label and an additional one for training the stop symbol. This can be used to speed up the computations.

There is no ground truth order in which the objects in an image must be predicted. Therefore simply matching predictions and ground truth labels as they appear would induce errors if the order does not match. For instance, if an image has objects A and B given in the ground truth labels in this order and the model predicts B, A, a direct matching would consider this prediction wrong, but in reality it is correct. As suggested by \cite{stewart2016end}, the Hungarian method is used as an efficient way to match ground truth labels to predictions using the distance between the coordinates as the matching cost.

A new loss function was derived based on \cite{stewart2016end} and \cite{redmon2016you}. The loss function $l$ is the sum of three components: the confidence (objectness) loss $l_o$, the coordinates (regression) loss $l_r$ and the class loss $l_c$.

For the following equations we consider the ground truth labels to be sorted such that the $i^{\text{th}}$ element of predictions is matched to the $i^{\text{th}}$ element of ground truth labels by the Hungarian method as described above.

The confidence score prediction is designed as a binary classification problem and we use softmax and cross-entropy loss for $l_o$.

The regression loss is a root mean squared error function that only penalises predictions where there is a real object. Define $\mathbf{g_o}$ to be the vector of ground truth confidence scores with elements $g_o(1), \dots, g_o(n) \in \{0,1\}$ where $n = G^2 * k$ is the number of predictions per image and $g_o(i) = 1$ if the prediction $i$ is an object, $0$ otherwise. Let $m = \sum_{i=1}^n{g_o(i)}$ be the number of real objects in the image, ${F_r}$ to be a $n \times r$ matrix, where $r$ is the number of coordinates required per object ($r=2$ for x, y coordinates and $r=4$ for bounding boxes) with elements $F_r(i, j)$ representing predicted coordinates. Similarly, let $G_r$ be a matrix of the same size with elements containing ground truth coordinates or zeros if there is no object. We now define the regression loss
\begin{equation}
  l_r = \sqrt {\frac{1}{m*r} \sum_{i=1}^n g_o(i) \sum_{j=1}^r (F_r(i, j) - G_r(i, j))^2}.
\end{equation}

Let $F_c$ be a $n \times C$ matrix ($C$ is the number of classes) with rows $\mathbf{F_c(i)}$ representing the probabilities of each class for object $i$ (output of softmax). Similarly let $G_c$ be a similar matrix representing the ground truth. The class loss is
\begin{equation}
l_c = \frac{1}{m} \sum_{i=1}^n g_o(i) H(\mathbf{F_c(i)}, \mathbf{G_c(i)}),
\end{equation}
where $H(\mathbf{a},\mathbf{b})$ is the cross-entropy between $\mathbf{a}$ and $\mathbf{b}$. Note this is the cross-entropy loss but it only penalises predictions where there is a real object (otherwise ground truth data does not exist for the class).

%

\section{Preprocessing}

The labels are pre-processed such that, for an image, we will have $G \times G \times k$ object labels in total, $k$ for each grid cell. If there are more than $k$ objects in a particular grid cell, we cap them at $k$ and drop the rest. Less than 10 objects are dropped in the whole crabs dataset for $k=8$. Objects with confidence $0$ are appended if there are fewer than $k$ labels.

Label coordinates are made relative to the centre of the grid cell. We say the top left corner of a grid cell is the point $(0, 0)$ and the bottom right corner is the point $(1, 1)$. The centre of the grid cell is at $(0.5, 0.5)$. This forces the LSTMs to predict objects that belong to their grid cell and makes the network easier to train. In the case of bounding boxes, the width and height are relative to the input image size, such that an object with width and height 1 covers the whole input image. This allows the network to predict objects that are larger than a grid cell.

The image pixel values are normalised to fall between -0.5 and 0.5.

\subsection{Crabs dataset} \label{subsection: crabs preprocessing}

\begin{table}[tbp]
\centering
\caption[Crabs dataset species counts]{Crabs species counts in the two locations (mosaics) and overall. The last column shows which species are in the \textbf{crabs-top3} dataset.}
\label{table:crabscount}
\begin{tabular}{lccc|c}
\hline
\textbf{Species}           & \multicolumn{1}{l}{\textbf{C0014}} & \multicolumn{1}{l}{\textbf{NBC}} & \multicolumn{1}{l|}{\textbf{Total}} & \multicolumn{1}{l}{\textbf{crabs-top3}} \\ \hline
Alvinocaridid              & 170                                          & 500                                        & 670                                 &                                         \\
Bathymodiolus japonicus    & 6,780                                         & 7,339                                       & 14,119                               & \checkmark                                       \\
Bathymodiolus platifrons   & 7,282                                         & 12,969                                      & 20,251                               & \checkmark                                       \\
Paralomis                  & 96                                           & 109                                        & 205                                 &                                         \\
Shinkaia crosnieri         & 3,536                                         & 7,160                                       & 10,696                               & \checkmark                                       \\
Thermosipho desbruyesi     & 12                                           & 6                                          & 18                                  &                                         \\ \hline
\textbf{Total individuals} & \textbf{17,876}                               & \textbf{28,083}                             & \textbf{45,959}                      & \textbf{45,066}
\end{tabular}
\end{table}

The crab dataset from \cite{thornton2016biometric} is made of two big mosaics C0014G\_2m\_2014 ($20320 \times 28448$ pixels) and NBC\_2m\_2014 ($20320 \times 20320$ pixels), each representing an imaging location. We slice the mosaics into $224 \times 224$ pixels images and randomly split them into training, cross-validation (dev) and test sets ($60\%-20\%-20\%$).

The labels consist of manually labelled x and y coordinates of each individual and its species.

The dataset is heavily unbalanced, as seen in Table \ref{table:crabscount}. This leads to hard to train models thus we split the dataset into two sub-datasets. We call \textbf{crabs} the dataset containing all images and \textbf{crabs-top3} the dataset consisting of the top 3 species by population count.

\begin{figure}[tpb]
  \centering
  \begin{subfigure}[b]{\textwidth}
    \centering
    \includegraphics[width=0.5\textwidth]{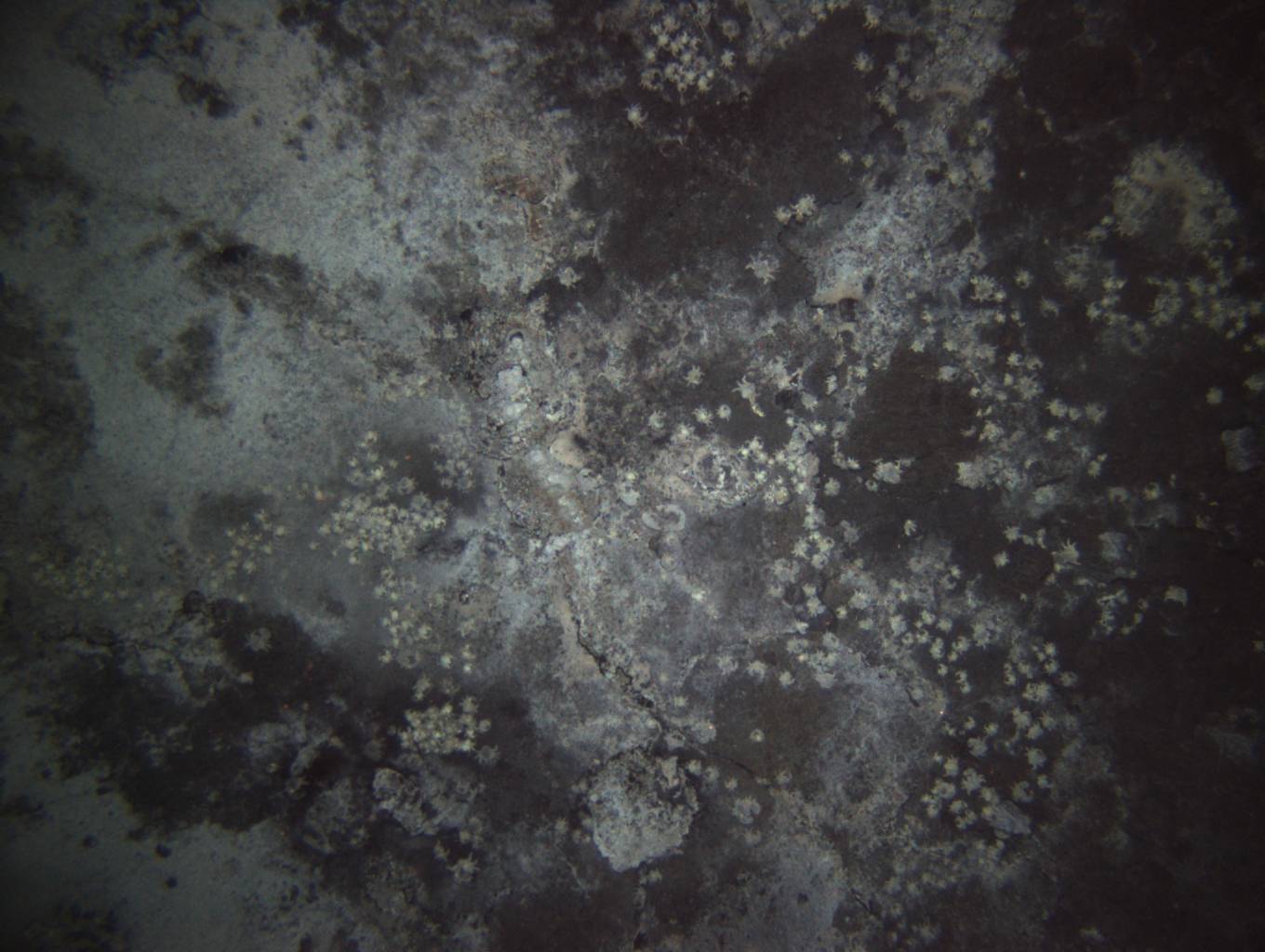}%
    ~
    \includegraphics[width=0.5\textwidth]{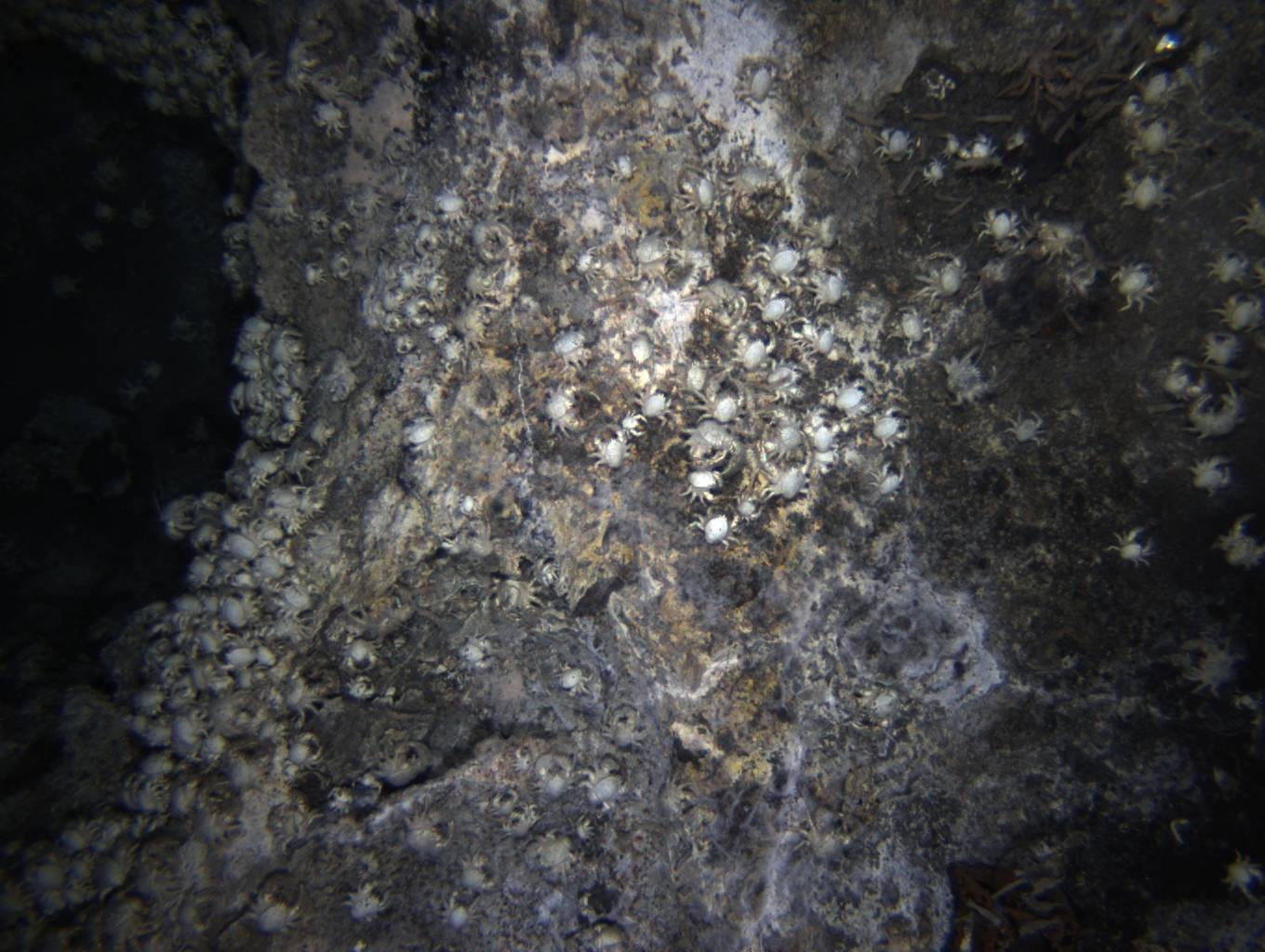}
    \caption{Examples of original images from the crabs dataset.}
  \end{subfigure}
  \begin{subfigure}[b]{\textwidth}
    \centering
    \includegraphics[width=0.5\textwidth]{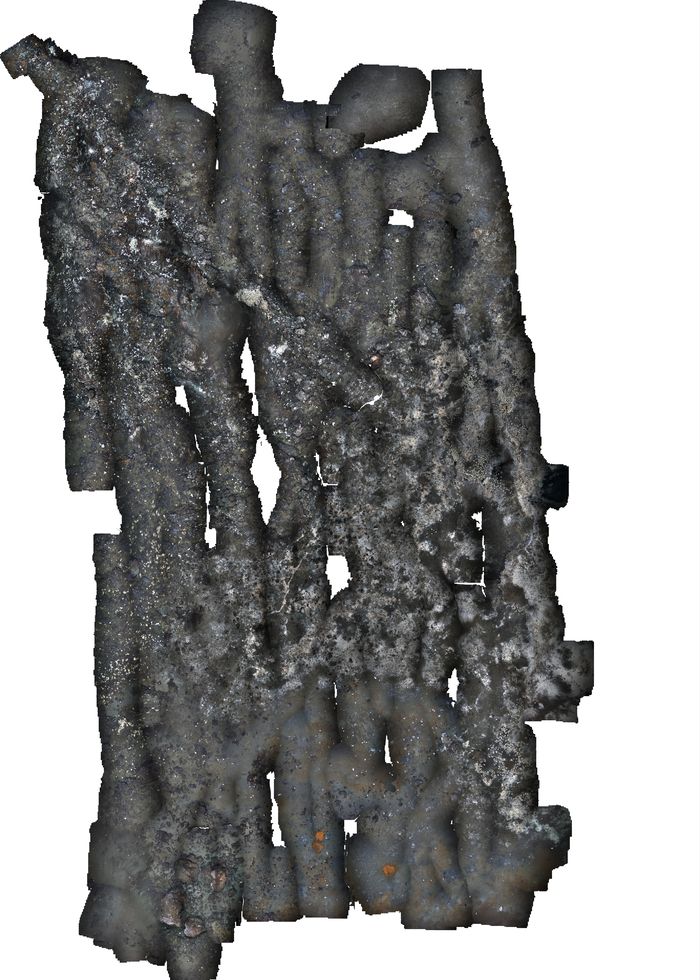}%
    ~
    \includegraphics[width=0.5\textwidth]{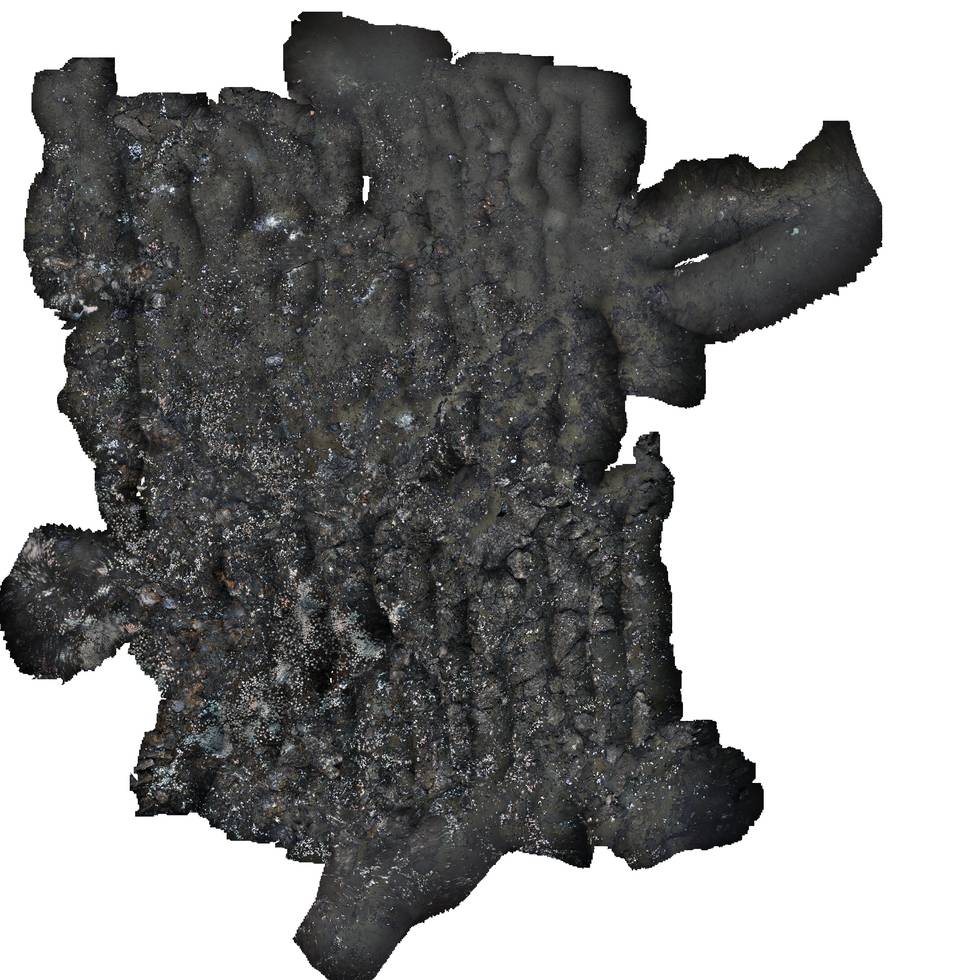}
    \caption{The two mosaics of the crabs dataset. Left: C0014G\_2m\_2014, right: NBC\_2m\_2014.}
  \end{subfigure}
  \begin{subfigure}[b]{\textwidth}
    \centering
    \includegraphics[width=0.4\textwidth]{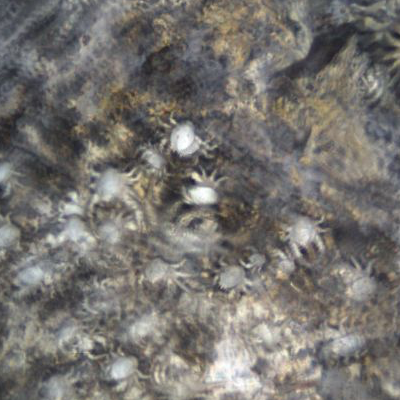}%
    ~
    \includegraphics[width=0.4\textwidth]{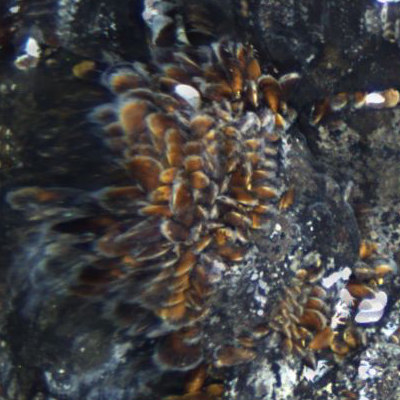}
    \caption{Two close-ups in the mosaics highlighting the effects of image quality loss, likely due to overlapping and image processing for creating the mosaic.}
  \end{subfigure}
  \caption[Overview of the crabs dataset]{Representation of the crabs dataset with example original images, the map mosaics from which the training data is sliced and highlights of image quality loss in the mosaics.}
  \label{fig:crabsdataset}
\end{figure}

The big mosaics provided suffer from image quality loss as they are composed of individual higher-quality images, overlapped and transformed to form a map-like mosaic. The effect can be observed in Figure \ref{fig:crabsdataset}. The original images are available but, unfortunately, there is no mapping between the labels on the mosaic and the original images, which are unlabelled.

Sequentially slicing the mosaics results in a total of 2608 $224 \times 224$ images. A dataset of such a small size is likely to make models easily overfit and hard to generalise. To address this issue we heavily augment the training set with random rotations, zooms, shears, and shifts and also use a different slicing technique where we cut an image centred on a crab. This results in a total of 45959 images (17,876 from C0014G and 28,083 from NBC) that have plenty of overlapping content.

\subsection{Sea Lions dataset} \label{subsection: sealion preprocessing}

The Steller Sea Lions dataset from the NOAA Kaggle competition \citep{noaakaggle} has many large aerial images. 948 for training and 18,636 for testing during the competition. Similarly to the crabs dataset, the dataset has x, y and class coordinates as labels, manually labeled.

The labels are given in two ways. Population counts per image are given in a CSV file. The x, y coordinates are given in a second pair of training images annotated with coloured dots, the colour denoting the class. The coordinates and classes can be extracted by subtracting the dotted and the pair non-dotted training image and running a blob detector on the result. However, this pre-processing step is a rather big source of errors in the training data used for our model. Population counts from the CSV and the extracted dotted images are shown in Table \ref{table: sea lion population}.

The given training images are large (sizes 3744x5616, 4992x3328 and 5616x3744). We slice them into 224x224 input images for training and testing. We split the given training set into a training and dev set (80\%-20\%). The resulting dataset has 21,683 images for training and 5243 for dev.

The competition goal, and evaluation criteria, is the final population counts per image rather than predicting where each individual is in the image.

\begin{table}[tbp]
\centering
\caption[Sea Lions dataset population counts]{Sea Lions dataset population counts, showing the differences between the given population counts and the generated dots. The dots have 16,955 dots that could not reliably be classified by colour (not shown).}
\label{table: sea lion population}
\begin{tabular}{llll}
                & \textbf{Label counts} & \textbf{Dot counts} & \textbf{$\Delta$} \\ \hline
adult\_males    & 5,392                  & 5,349                & 43             \\
subadult\_males & 4,345                  & 4,028                & 317            \\
adult\_females  & 37,537                 & 59,080               & 21,543          \\
juveniles       & 20,118                 & 20,895               & 777            \\
pups            & 16,285                 & 16,141               & 144            \\ \hline
\textbf{Totals} & \textbf{83,677}        & \textbf{105,493}     & \textbf{21,816}
\end{tabular}
\end{table}

\section{Training and evaluation}

A set of model configurations and hyper-parameters were chosen to train and evaluate our network with. The \ac{cnn} used to create a feature map is arguably the most important configuration choice.

We use the Inception-V1 (GoogLeNet) from \cite{szegedy2015going} up to the \emph{Mixed\_5c} layer, pre-trained on ImageNet by Google \footnote{Weights available at \url{https://github.com/tensorflow/models/tree/master/slim}}. The weights are not frozen during training.

The input image size is fixed at 224x224. A configurable parameter is the grid size $G$. The \ac{cnn} gives $G=7$ at the \emph{Mixed\_5c} layer. For $G=4$ we add an extra 4x4 stride 1 max pool layer.

The RNN has two parameters, the number of outputs per grid cell, $k$, and the number of LSTM layers used. We have 2 LSTM layers and $k=8$ for all experiments.

\subsection{Evaluation on crabs dataset}

The crabs dataset is prepared according to Section \ref{subsection: crabs preprocessing}. The model was trained on the \textbf{crabs} dataset (all 6 classes) for 15 epochs using standard data augmentation methods (random rotations, shears, zooms and shifts) to prevent overfitting. We obtain a \ac{mAP} of 24.14\% for a 4x4 grid and 19.32\% for a 7x7 grid on test, however there is a large difference between the top 3 and bottom 3 classes as seen in Table \ref{table: crabs eval all} (Sliced sequentially). This is due to the dataset being heavily unbalanced.

A similar training scenario for the \textbf{crabs-top3} dataset yields 37.5\% mAP using a 4x4 grid and 27.37\% mAP for the 7x7 grid. Results are shown in Table \ref{table: crabs eval top 3} (Sliced sequentially).

As stated in Section \ref{subsection: crabs preprocessing}, sequentially slicing the two mosaics results in a relatively tiny dataset for deep models. We therefore slice the mosaics to produce one image for each object by cutting out around the object coordinates. This creates many overlapping images which can be seen as a data augmentation technique. We pick one mosaic for training and one for testing. The dev set is sliced around objects and the test set is sequentially sliced. We trained the models for 15 epochs. We ran this experiment four times, using each mosaic as training in turn and using all or only the top 3 classes. All experiments overfit the training set and perform the poorest on the test set. Results are in Tables \ref{table: crabs eval all} and \ref{table: crabs eval top 3} (Sliced around object).

\begin{figure}
	\centering
	\includegraphics[width=0.5\textwidth]{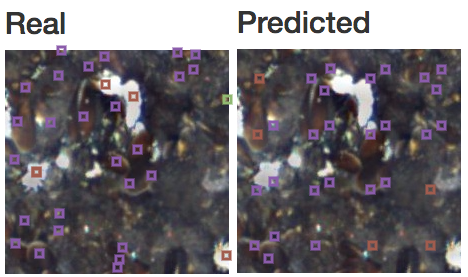}~
	\includegraphics[width=0.5\textwidth]{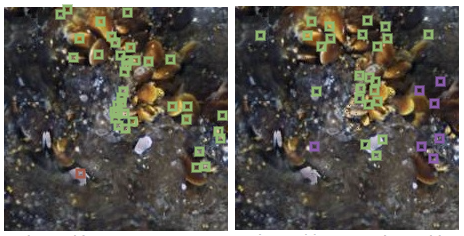}
	\includegraphics[width=0.5\textwidth]{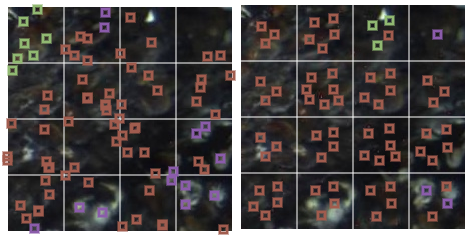}~
	\includegraphics[width=0.5\textwidth]{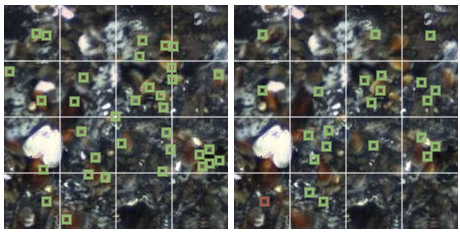}
	\includegraphics[width=0.5\textwidth]{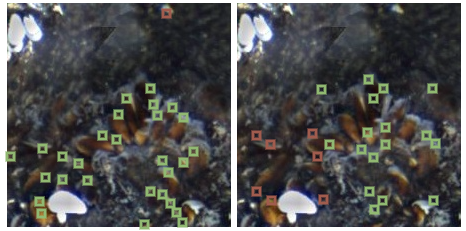}~
	\includegraphics[width=0.5\textwidth]{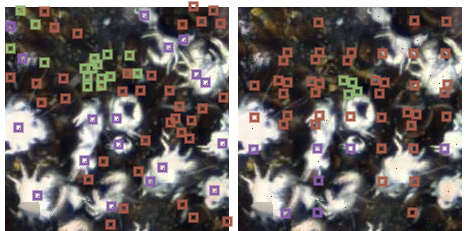}
	\caption[Example predictions on the crabs dataset]{Example predictions on the crabs dataset. The grid is rendered in the middle row for reference. Classes (species) are colour coded. For each image pair, ground truth on left and prediction on right. Confidence threshold 0.5.}
	\label{fig:crab-pics}
\end{figure}

Example predictions are shown in Figure \ref{fig:crab-pics}, where we can observe the model does large localisation errors but at the same time it predicts good population densities. The locations of predicted crabs tend to form a circle centred near the middle of the grid cells.

\begin{table}[tbp]
\centering
\caption[Evaluation on the crabs dataset (all classes)]{Model evaluation on the crabs dataset with all classes used. 7x7 and 4x4 denote the grid sizes used and slicing states the method of slicing the mosaics into training images. For slicing around objects, the training set is based on the NBC mosaic, the dev set on C0014G and the test set is sequentially sliced. Class values show AP \%, mAP in \%.}
\label{table: crabs eval all}
\resizebox{\textwidth}{!}{%
\begin{tabular}{l|ccc|ccc|ccc|ccc|}
\cline{2-13}
                                               & \multicolumn{6}{c|}{Sliced sequentially}                                                                & \multicolumn{6}{c|}{Sliced around objects}                                                                  \\ \cline{2-13}
                                               & \multicolumn{3}{c|}{4x4}                         & \multicolumn{3}{c|}{7x7}                         & \multicolumn{3}{c|}{4x4}                         & \multicolumn{3}{c|}{7x7}                         \\ \cline{2-13}
                                               & Train          & Dev            & Test           & Train          & Dev            & Test           & Train          & Dev            & Test           & Train          & Dev            & Test           \\ \hline
\multicolumn{1}{|l|}{Bathymodiolus japonicus}  & 61.12        & 36.81        & 38.50        & 12.41        & 10.02        & 10.38        & 71.00        & 41.63        & 22.46        & 46.65        & 9.27         & 9.26         \\
\multicolumn{1}{|l|}{Thermosipho desbruyesi}   & 9.17         & 9.29         & 9.19         & 46.74        & 28.91        & 32.53        & 26.08        & 9.11         & 9.12         & 55.51        & 25.87        & 18.08        \\
\multicolumn{1}{|l|}{Bathymodiolus platifrons} & 60.81        & 40.81        & 33.27        & 52.34        & 28.65        & 31.01        & 75.59        & 36.14        & 20.59        & 61.33        & 30.38        & 17.89        \\
\multicolumn{1}{|l|}{Paralomis}                & 12.49        & 10.02        & 9.59         & 11.74        & 9.48         & 9.54         & 14.89        & 9.30         & 9.72         & 20.72        & 9.15         & 9.24         \\
\multicolumn{1}{|l|}{Alvinocaridid}            & 11.98        & 10.30        & 11.04        & 50.82        & 30.34        & 23.37        & 46.43        & 9.81         & 9.71         & 60.18        & 28.05        & 16.38        \\
\multicolumn{1}{|l|}{Shinkaia crosnieri}       & 56.61        & 39.71        & 43.27        & 9.12         & 9.10         & 9.11         & 69.49        & 32.71        & 22.86        & 44.01        & 9.10         & 9.10         \\ \hline
\multicolumn{1}{|l|}{\textbf{mAP}}             & \textbf{35.36} & \textbf{24.49} & \textbf{24.14} & \textbf{30.53} & \textbf{19.42} & \textbf{19.32} & \textbf{50.58} & \textbf{23.12} & \textbf{15.74} & \textbf{48.07} & \textbf{18.64} & \textbf{13.33} \\ \hline
\end{tabular}%
}
\end{table}

\begin{table}[tbp]
\centering
\caption[Evaluation on the crabs-top3 dataset (top 3 classes)]{Model evaluation on the \textbf{crabs-top3} dataset (only top 3 classes used). 7x7 and 4x4 denote the grid sizes used and slicing states the method of slicing the mosaics into training images. For slicing around objects, the training set is based on the NBC mosaic, the dev set on C0014G and the test set is sequentially sliced. Class values show AP \%, mAP in \%.}
\label{table: crabs eval top 3}
\resizebox{\textwidth}{!}{%
\begin{tabular}{l|ccc|ccc|ccc|ccc|}
\cline{2-13}
                                               & \multicolumn{6}{c|}{Sliced sequentially}                                                                & \multicolumn{6}{c|}{Sliced around objects}                                                                  \\ \cline{2-13}
                                               & \multicolumn{3}{c|}{4x4}                         & \multicolumn{3}{c|}{7x7}                         & \multicolumn{3}{c|}{4x4}                         & \multicolumn{3}{c|}{7x7}                         \\ \cline{2-13}
                                               & Train          & Dev            & Test           & Train          & Dev            & Test           & Train          & Dev            & Test           & Train          & Dev            & Test           \\ \hline
\multicolumn{1}{|l|}{Bathymodiolus japonicus}  & 64.02          & 36.16          & 38.49          & 51.04          & 27.35          & 29.26          & 71.55          & 41.45          & 23.72          & 61.61          & 30.15          & 19.00          \\
\multicolumn{1}{|l|}{Bathymodiolus platifrons} & 63.65          & 38.23          & 32.94          & 47.78          & 29.58          & 24.22          & 75.61          & 36.25          & 21.01          & 60.14          & 28.47          & 17.89          \\
\multicolumn{1}{|l|}{Shinkaia crosnieri}       & 59.66          & 35.71          & 40.61          & 41.55          & 25.66          & 28.63          & 68.97          & 32.60          & 24.06          & 55.29          & 24.15          & 17.26          \\ \hline
\multicolumn{1}{|l|}{\textbf{mAP}}             & \textbf{62.44} & \textbf{36.70} & \textbf{37.35} & \textbf{46.79} & \textbf{27.53} & \textbf{27.37} & \textbf{72.04} & \textbf{36.77} & \textbf{22.93} & \textbf{59.01} & \textbf{27.59} & \textbf{18.05} \\ \hline
\end{tabular}%
}
\end{table}

\subsection{Evaluation on VOC12 dataset}

Two models were trained on the VOC dataset, one using a 4x4 grid and one using a 7x7 grid. The resulting mAPs on the dev test are 29.93\% and 27.34\%, respectively, which are below the state of the art but the model parameters are not fine-tuned for this dataset. We focus mainly on small objects that fit into a grid cell. The training was done with no data augmentation. Table \ref{table:eval-voc} shows per-class average precision (AP) scores and mAP results on train and dev sets.

Large objects that span across many grid cells are predicted more than once, but only considered correct in the cell containing the centre of the ground truth label. This raises the level of false positives at test time and can negatively influence training. For instance, if an object is in the middle between two grid cells, both are equally qualified to make that prediction, but only one will contain the object and during training this situation will be regarded as an error in the other grid cell.

\begin{figure}
	\centering
	\includegraphics[width=0.5\textwidth]{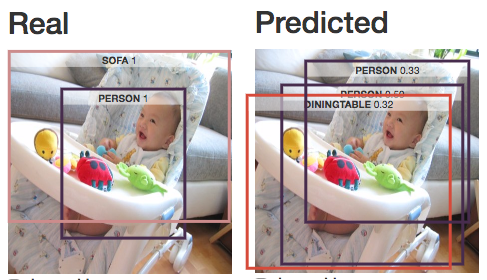}~
	\includegraphics[width=0.5\textwidth]{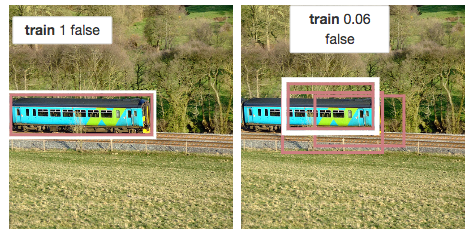}
	\includegraphics[width=0.5\textwidth]{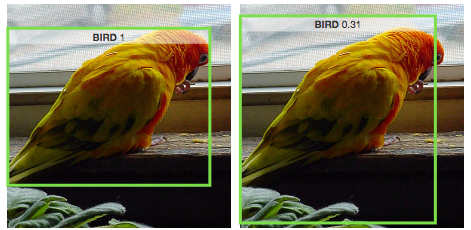}~
	\includegraphics[width=0.5\textwidth]{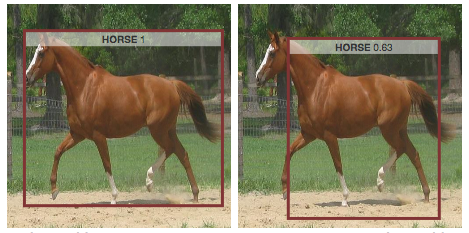}
	\includegraphics[width=0.5\textwidth]{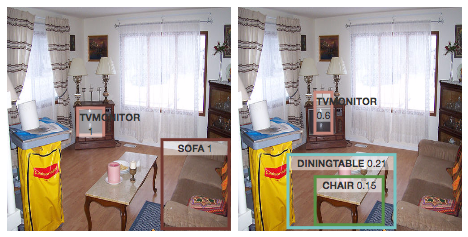}~
	\includegraphics[width=0.5\textwidth]{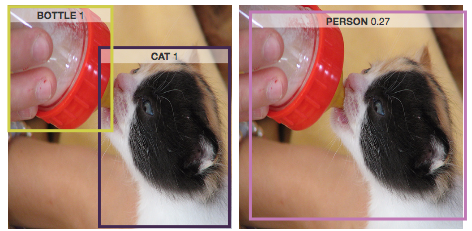}
	\caption[Example predictions on the VOC12 dataset]{Example predictions on the VOC12 dataset. For each image pair, ground truth on left and prediction on right. Confidence score shown next to object class name.}
	\label{fig:predictions-voc}
\end{figure}

Example predictions are shown in Figure \ref{fig:predictions-voc}

\begin{table}[tbp]
\centering
\caption[Evaluation on VOC12]{Evaluation results on the VOC12 detection dataset. Class results are Average Precision (AP). 4x4 and 7x7 denote the grid size of our model configuration. Input images resized to 224x224 pixels. No data augmentation was used and no fine-tuning of the model. Values in \%.}
\label{table:eval-voc}
\begin{tabular}{lllll}
             & Train 7x7        & Train 4x4        & Dev 7x7         & Dev 4x4 \\ \hline
aeroplane    & 51.66          & 52.74          & 36.69          & 34.29  \\
bicycle      & 55.19          & 52.00          & 22.46          & 23.37  \\
bird         & 42.37          & 52.01          & 29.04          & 33.65  \\
boat         & 45.87          & 50.67          & 24.00          & 27.47  \\
bottle       & 51.56          & 46.62          & 14.79          & 14.38  \\
bus          & 69.35          & 56.27          & 38.57          & 35.07  \\
car          & 52.61          & 56.92          & 27.99          & 32.24  \\
cat          & 47.69          & 61.27          & 31.56          & 38.95  \\
chair        & 39.57          & 52.36          & 18.57          & 22.95  \\
cow          & 57.38          & 56.83          & 25.57          & 31.30  \\
diningtable  & 54.00          & 69.85          & 20.41          & 24.91  \\
dog          & 51.58          & 62.07          & 31.08          & 35.35  \\
horse        & 49.49          & 61.69          & 23.50          & 32.43  \\
motorbike    & 55.91          & 58.58          & 29.69          & 28.17  \\
person       & 54.09          & 65.14          & 40.60          & 47.76  \\
pottedplant  & 40.74          & 49.22          & 18.11          & 21.33  \\
sheep        & 55.27          & 60.86          & 35.92          & 36.32  \\
sofa         & 52.88          & 65.97          & 16.83          & 20.28  \\
train        & 47.68          & 47.79          & 30.18          & 32.39  \\
tvmonitor    & 51.37          & 53.66          & 31.16          & 25.95  \\ \hline
\textbf{mAP} & \textbf{51.31} & \textbf{56.63} & \textbf{27.34} & \textbf{29.93}
\end{tabular}
\end{table}

\subsection{Evaluation on Steller Sea Lions Dataset}

The aim of the NOAA Fisheries Kaggle competition is to predict population counts, not to detect where objects are. The evaluation metric for the competition is mean column-wise root mean squared error (RMSE). For N images and K classes, it is:
\begin{equation}
	\frac{1}{K} \sum_i^K \sqrt{ \frac{1}{N} \sum_j^N  (y_{ij}-\hat{y}_{ij})^2 },
\end{equation}
where the labels $y$ and predictions $\hat{y}$ are population counts.

We trained two models, grid 7x7 and grid 4x4. On the dev set we achieve 31.40\% and 38.33\%, respectively. Training was done for 30 epochs. Mean Column-Wise RMSE at 0.5 confidence threshold for 7x7 and 4x4 are 1.68 and 1.52, respectively. Note this is on the sliced dev set, not on the large images from the original dataset. The values will likely be higher on the larger images. Results illustrated in Table \ref{table: sealioneval}. At the time of writing, the top position in the competition leaderboard has 11.33729 RMSE on the official test set.

\begin{table}[tbp]
\centering
\caption[Evaluation results on the Steller Sea Lions dataset]{Evaluation results on the Steller Sea Lions dataset. Class results are Average Precision (AP) \%. 4x4 and 7x7 denote the grid size of our model configuration. Dataset images sequentially sliced in 224x224 input images. mAP in \%. RMSE is Mean Column-Wise RMSE at 0.5 confidence threshold.}
\label{table: sealioneval}
\begin{tabular}{lcccc}
               & \textbf{7x7 train} & \textbf{7x7 dev} & \textbf{4x4 train} & \textbf{4x4 dev} \\ \hline
adult females  & 35.68              & 33.00            & 45.03              & 41.92            \\
adult males    & 49.24              & 36.77            & 57.81              & 45.86            \\
juveniles      & 39.91              & 31.87            & 47.52              & 38.86            \\
pups           & 36.38              & 32.36            & 43.27              & 37.98            \\
subadult males & 36.01              & 23.01            & 48.67              & 27.04            \\ \hline
\textbf{mAP}   & \textbf{39.44}     & \textbf{31.40}   & \textbf{48.46}     & \textbf{38.33} \\
\textbf{RMSE}   & \textbf{1.69}     & \textbf{1.68}   & \textbf{1.49}     & \textbf{1.52}
\end{tabular}
\end{table}

\section{Implementation}

The model was developed with TensorFlow \citep{tensorflow2015-whitepaper}. I used parts of TensorBox\footnote{https://github.com/TensorBox/TensorBox} (TensorFlow implementation of \citep{stewart2016end}) as a reference point. My implementation is an object detection network as opposed to only object localisation like TensorBox.

I implemented image and label readers for VOC, sea lions and crabs datasets along with tools used for slicing images and labels. I implemented an evaluation, prediction and visualisation tool. The predictions are stored in an sqlite databases and can be later used for visualisation and further analysis.

\begin{figure}
	\centering
	\includegraphics[width=\textwidth]{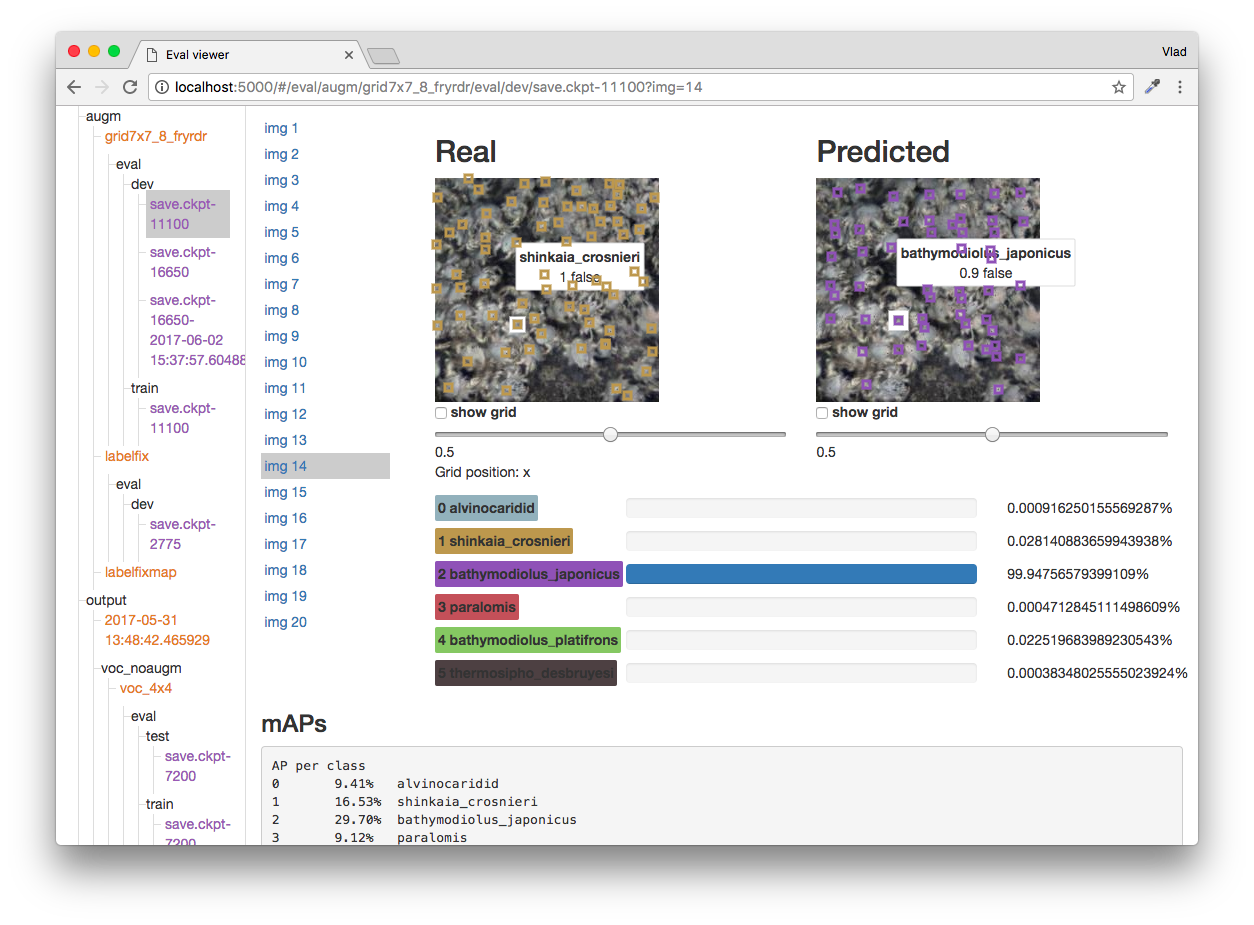}
	\caption[Screenshot of the visualisation tool]{Screenshot of the visualisation tool built to aid the development of this model. The predictions seen in the page are all misclassified. Highlighting shows matching pair of ground truth and prediction.}
	\label{fig:vis-screenshot_crabs}
\end{figure}

The visualisation tool allows viewing all trained models in a directory tree along with relevant evaluation results on test, dev and training sets. The visualisation tool is a python (with flask) web server and web page that shows evaluation metrics, class counts, images and predictions on the images and precision-recall plots. A screenshot is shown in Figure \ref{fig:vis-screenshot_crabs}.

\bibliography{nips_2018}
\bibliographystyle{nips_2018}

\end{document}